\documentclass[11pt]{article}

\usepackage{fontawesome5}
\usepackage[utf8]{inputenc}
\usepackage[T1]{fontenc}
\usepackage{newtxtext,newtxmath}
\usepackage[margin=1in]{geometry}
\usepackage{amsmath,amsfonts}
\usepackage{booktabs}
\usepackage{xcolor}
\usepackage{algorithm}
\usepackage{algpseudocode}
\usepackage{hyperref}
\usepackage{cleveref}
\usepackage{titlesec}
\usepackage{enumitem}
\usepackage{graphicx}
\usepackage{tikz}
\usepackage[numbers,sort&compress]{natbib}
\usepackage{fancyhdr}
\usetikzlibrary{shapes.geometric, arrows.meta, positioning, fit, backgrounds}

\definecolor{iclr_blue}{RGB}{42, 82, 120}
\definecolor{novo_gold}{RGB}{184, 134, 11}

\titleformat{\section}{\Large\bfseries\color{iclr_blue}}{\thesection}{1em}{}
\titleformat{\subsection}{\large\bfseries}{\thesubsection}{1em}{}
\titleformat{\subsubsection}{\normalsize\bfseries}{\thesubsubsection}{1em}{}

\renewcommand{\headrulewidth}{0.5pt}
\renewcommand{\headrule}{\hbox to\headwidth{\color{novo_gold}\leaders\hrule height \headrulewidth\hfill}}
\renewcommand{\footrulewidth}{0.5pt}
\renewcommand{\footrule}{\hbox to\headwidth{\color{novo_gold}\leaders\hrule height \footrulewidth\hfill}}

\fancypagestyle{firstpage}{
  \fancyhf{}
  \lhead{\raisebox{-4pt}[0pt][0pt]{\textcolor{novo_gold}{\textbf{Novo Ordo for AI}}}}
  \rhead{\raisebox{-4pt}[0pt][0pt]{\textcolor{gray}{2026-06-24}}}
  \rfoot{\textcolor{gray}{\thepage}}
  \renewcommand{\headrulewidth}{0.5pt}
  \renewcommand{\headrule}{\hbox to\headwidth{\color{novo_gold}\leaders\hrule height \headrulewidth\hfill}}
  \renewcommand{\footrulewidth}{0.5pt}
  \renewcommand{\footrule}{\hbox to\headwidth{\color{novo_gold}\leaders\hrule height \footrulewidth\hfill}}
}

\makeatletter
\def\maketitle{%
  \newpage
  \null
  \vskip 2em%
  \begin{center}%
  \includegraphics[width=1.5cm]{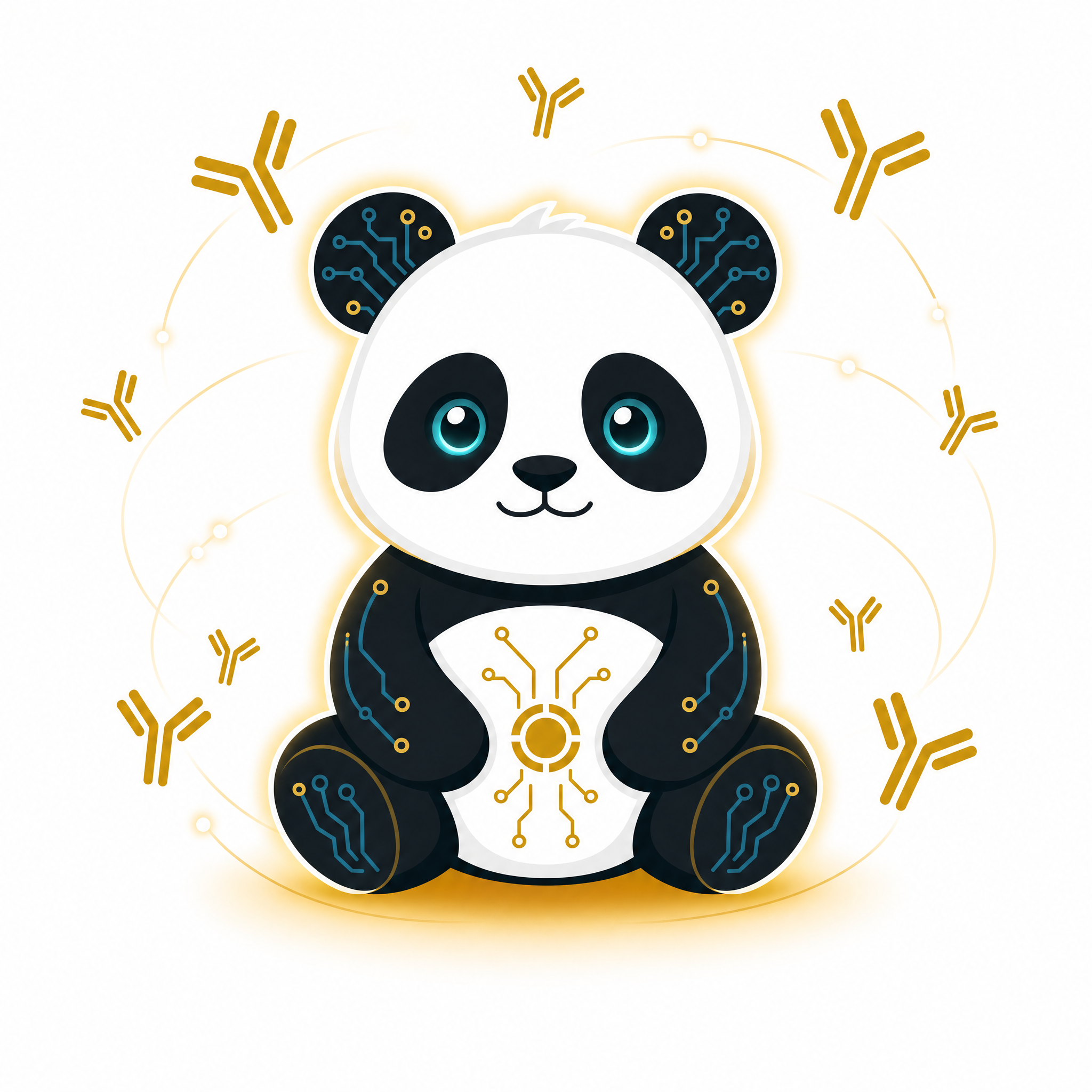}\\[0.5em]  % 吉祥物图标
  \let \footnote \thanks
    {\LARGE\bfseries\@title \par}%
    \vskip 1.0em%
    {\large
      \lineskip .5em%
      \begin{tabular}[t]{c}%
        \@author
      \end{tabular}\par}%
  \end{center}%
  \par
  \vskip 1.5em
  \thispagestyle{firstpage}%
}
\makeatother

\title{Agent-Native Immune System:\\Architecture, Taxonomy, and Engineering}
\author{
Bo Shen, Lifeng Chang, Tianyuan Wei, Yunpeng Li, \\
Feng Shi, Yichen Han, Peijie Gao, Shiyi Kuang, Xin Chang, Dehui Li \vspace{0.3em} \\
Novo Ordo for AI
}

\begin{document}

\maketitle

% ============================================================
% ABSTRACT
% ============================================================
\begin{abstract}
\noindent
The transition from static chat bots to autonomous agents---equipped with persistent memory, tool-use protocols, and multi-agent collaboration---has fundamentally expanded the AI threat landscape. Current defense mechanisms, such as perimeter security and training-time alignment, remain external to the agent's active reasoning loop. Consequently, they fall short: a fully aligned agent remains highly vulnerable to runtime hijacking via memory poisoning, tool-chain manipulation, or multi-agent protocol attacks.

To address this critical gap, we introduce the \textbf{Agent-Native Immune System (ANIS)}, the first biologically inspired, endogenous defense architecture embedded directly within the agent's cognitive loop. Our framework presents four primary contributions. \textbf{First}, we design a six-layer \textit{Immune Tower} (L0--L5), distinctly incorporating \textit{Barrier Immunity} (L1) as a non-cognitive, physical-and-logical isolation layer. \textbf{Second}, we establish a unified taxonomy of \textit{Agent Viruses} and \textit{Agent Vaccines}, formalizing the critical distinction between superficial \textit{non-parametric} defenses and robust \textit{parametric} vaccines. \textbf{Third}, we conceptualize the \textbf{Harness Triad}---Meta, Self, and Auto---a self-monitoring, meta-cognitive automation backbone that drives \textit{Continual Immune Learning (CIL)}, enabling vaccines to dynamically adapt to novel threats. \textbf{Finally}, we establish a rigorous theoretical demarcation between model alignment and agent immunity: while alignment provides a static ``constitutional'' value foundation during training, ANIS serves as the dynamic ``law enforcement'' mechanism during runtime.

We conclude by framing open challenges for the field, including immune protocol standardization, novel evaluation metrics such as the \textit{Autoimmunity Rate} (false-positive intervention rate), and the co-evolutionary dynamics between pathogens and vaccines within collective intelligence ecosystems.
\end{abstract}

% ============================================================
% 1. INTRODUCTION
% ============================================================
\section{Introduction}
\label{sec:intro}

\textbf{From Reactive Tools to Proactive Agents.} The evolution of large language models traces a clear trajectory from passive responders to active agents. Early systems excelled at completion (GPT-3). The chat paradigm (ChatGPT) introduced conversational alignment. Reasoning models (OpenAI o1, DeepSeek-R1) enabled step-by-step deliberation. Perception-and-action models (Claude Sonnet 3.5, Gemini 3 Pro) bridged digital and physical environments. The frontier is now collaborative intelligence (Claude Opus 4.6, Kimi Agent Swarm), where agents coordinate in persistent collectives.
This capability expansion reflects a deeper shift in engineering paradigms. Prompt engineering optimized static text inputs. Context engineering expanded the optimizable surface to retrieved documents and memory buffers. Intent engineering emerged as a critical layer: encoding enterprise goals, value hierarchies, and trade-off priorities into the agent's decision substrate, addressing the strategic deficit where "the agent sees all the right information but still optimizes the wrong objective" \citep{vishnyakova2026intent}. Harness engineering \citep{lee2026meta,lou2026autoharness,zhang2026selfharness} treats the entire surrounding system as a unified object of optimization. Loop engineering closes the feedback loop: systems that observe, adapt, and improve without human intervention.

This progression mirrors the shift from reactive to proactive AI. Prompt engineering serves user-initiated queries. Context engineering supports stateful sessions. Harness engineering enables autonomous action. Loop engineering creates self-improving agents that persist, adapt, and evolve---virtual digital employees. With this autonomy comes an isomorphic expansion of the attack surface. Each new capability introduces a new vulnerability class. Tool-use exposes agents to adversarial tool metadata and supply-chain attacks. Persistent memory creates a persistent attack surface: a single poisoned entry can bias decisions indefinitely. Multi-agent collaboration introduces protocol-level manipulation. Local-first gateways like OpenClaw bridge sandboxed cloud APIs and real-world system access. The agent's cognitive state---goals, memories, tool bindings, peer relationships---is under continuous threat.

\begin{figure*}[t]
\centering
\scalebox{0.92}{%
\begin{tikzpicture}[
    milestone/.style={rectangle, rounded corners=2pt, minimum width=1.8cm, minimum height=0.6cm, align=center, font=\scriptsize, draw, thick},
    model/.style={milestone, fill=blue!10, draw=blue!40, text=blue!60!black},
    agent/.style={milestone, fill=novo_gold!8, draw=novo_gold!60, text=novo_gold!80!black},
    goal/.style={milestone, fill=green!8, draw=green!40, text=green!50!black, minimum width=2.6cm, minimum height=0.5cm},
    year/.style={font=\bfseries\footnotesize},
    arrow/.style={-{Stealth[length=1.5mm]}, thick}
]

% 上方：Foundation Model 能力演进
\node[model] (m1) at (0,1.4) {Completion\\(GPT-3)};
\node[model] (m2) at (2,1.4) {Chat\\(ChatGPT)};
\node[model] (m3) at (4,1.4) {Tool-use\\(GPT-4)};
\node[model] (m4) at (6,1.4) {ReAct\\(Sonnet 3.5)};
\node[model] (m5) at (8,1.4) {Reasoning\\(o1/R1)};
\node[model] (m6) at (10,1.4) {Collaboration\\(Opus 4.6)};

\draw[arrow, blue!40] (m1) -- (m2);
\draw[arrow, blue!40] (m2) -- (m3);
\draw[arrow, blue!40] (m3) -- (m4);
\draw[arrow, blue!40] (m4) -- (m5);
\draw[arrow, blue!40] (m5) -- (m6);

\node[font=\bfseries\small, blue!60] at (5,2.2) {Foundation Model Capabilities};

% 下方：Agent Engineering 范式演进
\node[agent] (a1) at (0,-1.4) {Prompt\\Eng.};
\node[agent] (a2) at (2,-1.4) {Context\\Eng.};
\node[agent] (a3) at (4,-1.4) {Intent\\Eng.};
\node[agent] (a4) at (6,-1.4) {Harness\\Eng.};
\node[agent] (a5) at (8,-1.4) {Loop\\Eng.};
\node[agent] (a6) at (10,-1.4) {Immune\\Eng.};

\draw[arrow, novo_gold!60] (a1) -- (a2);
\draw[arrow, novo_gold!60] (a2) -- (a3);
\draw[arrow, novo_gold!60] (a3) -- (a4);
\draw[arrow, novo_gold!60] (a4) -- (a5);
\draw[arrow, novo_gold!60] (a5) -- (a6);

\node[font=\bfseries\small, novo_gold!80] at (5,-2.2) {Agent Engineering Paradigms};

% 时间轴
\draw[thick, gray] (-0.5,0) -- (10.5,0);
\foreach \x/\year in {0/2020, 2/2022, 4/2023, 6/2024, 8/2025, 10/2026} {
    \draw[thick, gray] (\x,-0.1) -- (\x,0.1);
    \node[year, below] at (\x,-0.2) {\year};
}

% 右侧汇聚：灰色线无箭头，自然汇入交汇点
\coordinate (hub) at (11.5, 0);
\draw[thick, gray!60] (m6.east) -- ++(0.5,0) |- (hub);
\draw[thick, gray!60] (a6.east) -- ++(0.5,0) |- (hub);

% 从交汇点向右的绿色主干（无箭头）
\coordinate (branch) at (12.8, 0);
\draw[thick, green!40] (hub) -- (branch);

% ANIS Goals 标签
\node[font=\bfseries\small, green!50!black] at (14.5, 1.6) {ANIS Goals};

% 四个目标模块（带fontawesome5图标）
\node[goal] (g1) at (14.5, 1.0) {\faLock\ \textbf{SECURE}};
\node[goal] (g2) at (14.5, 0.4) {\faHeartbeat\ \textbf{HEALTHY}};
\node[goal] (g3) at (14.5, -0.2) {\faListUl\ \textbf{ORDERLY}};
\node[goal] (g4) at (14.5, -0.8) {\faSync\ \textbf{EVOLVING}};

% 从branch分发到四个目标（带箭头）
\draw[arrow, green!40] (branch) |- (g1.west);
\draw[arrow, green!40] (branch) |- (g2.west);
\draw[arrow, green!40] (branch) |- (g3.west);
\draw[arrow, green!40] (branch) |- (g4.west);

\end{tikzpicture}%
}
\caption{The co-evolution of foundation model capabilities and agent engineering paradigms, converging toward the four goals of ANIS: Secure, Healthy, Orderly, and Evolving.}
\label{fig:evolution}
\end{figure*}
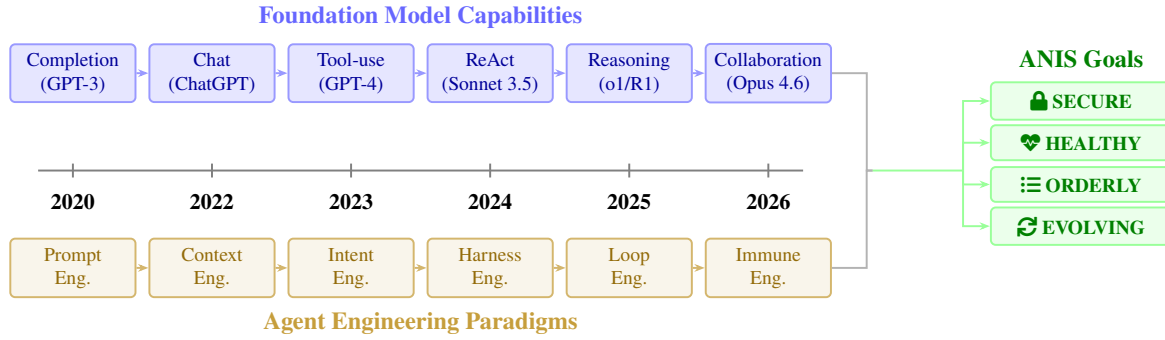

As illustrated in \Cref{fig:evolution}, the co-evolution of model capabilities and agent engineering paradigms motivates the need for a unified immune framework. We propose the next stage: immune engineering, which ensures that self-improving agents remain secure, healthy, and orderly throughout their operational lifetime, while continuously evolving their defensive capabilities.

\textbf{Existing defenses fall short.} Perimeter safeguards intercept attacks before they reach the agent. Training-time alignment embeds human values into model weights, but it is static. It cannot respond to novel runtime attacks that were unseen during training. Concrete attacks have demonstrated this vulnerability at every layer. \citet{chen2024agentpoison} show that memory poisoning installs persistent backdoors. \citet{memmorph2026} demonstrate that three crafted memory records can hijack tool selection with over 70\% attack success. \citet{li2025mcpinspect} expose cross-entity risks in the MCP ecosystem, where adversarial tool metadata shapes reasoning without code-level vulnerabilities. \citet{zhang2026msb} benchmark MCP attacks at scale, identifying hundreds of vulnerable servers. \citet{weckbecker2026thought} introduce ``thought viruses'' that propagate viral misalignment across agent swarms. \citet{hu2026collusion} demonstrate open-channel multi-agent collusion for belief manipulation. These attacks bypass perimeter defenses by operating inside the agent's reasoning, memory, or inter-agent protocols.

What is missing is an \textit{endogenous} system that targets at the agent's \textbf{security} (protection from external threats), \textbf{health} (preservation of internal integrity and goal stability), \textbf{order} (governance of multi-agent interactions), and \textbf{evolution} (continuous adaptation and improvement of defensive capabilities). Biological organisms solved this problem through a multi-tiered immune system capable of distinguishing self from non-self, mounting rapid innate responses, generating adaptive antibodies, and retaining immunological memory.

We formalize the \textbf{Agent-Native Immune System (ANIS)}. Unlike prior analogies between computer security and immunology (e.g., \citealp{forrest1997sense}), which targeted static software or network intrusion detection, ANIS addresses a continuously reasoning, goal-directed entity. We advance beyond metaphor to engineering: we define precise taxonomies, specify parametric and non-parametric vaccine mechanisms, and introduce the \textbf{Harness Triad} that operationalizes adaptive immunity in a deployable framework.

\textbf{Contributions.} We make the following contributions:
\begin{enumerate}[leftmargin=1.5em, itemsep=0.2em]
    \item \textbf{The Immune Tower (L0--L5):} A six-layer integer-indexed architecture mapping biological immunity to agent engineering, with explicit barrier immunity (L1) as a non-cognitive isolation layer.
    \item \textbf{Unified Taxonomy of Viruses and Vaccines:} The first comprehensive ontology of agent pathogens and defenses, distinguishing non-parametric vaccines (rules, prompts) from parametric vaccines (steering vectors, LoRA adapters, defensive embeddings).
    \item \textbf{The Harness Triad and Continual Immune Learning (CIL):} We redirect three paradigms from harness engineering---meta-level search \citep{lee2026meta}, automatic synthesis \citep{lou2026autoharness}, and self-improvement \citep{zhang2026selfharness}---toward immune defense, forming an engineering framework for adaptive, self-improving immunity.
    \item \textbf{Security--Health--Order--Evolution Unification:} We rigorously differentiate ANIS from model alignment, arguing that alignment provides constitutional values while ANIS provides runtime law enforcement, and that together they constitute complementary pillars of robust agentic development.
\end{enumerate}

% ============================================================
% 2. BACKGROUND & RELATED WORK
% ============================================================
\section{Background and Related Work}
\label{sec:background}

\subsection{The Agent Engineering Evolution}

The agent stack has evolved through five distinct engineering paradigms. Prompt engineering (Brown et al., 2020; Liu et al., 2023) optimized static text inputs to elicit better outputs from fixed models. Context engineering (Liu, 2023; Mallen et al., 2023) expanded the optimizable surface to include retrieved documents, memory buffers, and dynamically constructed inputs. Intent engineering \citep{vishnyakova2026intent} addresses a deeper question: not just what the agent sees (context), but what it values. It encodes enterprise goals, value hierarchies, and trade-off priorities---such as trust over speed over cost---into the agent's decision substrate, solving the strategic deficit where agents optimize locally correct but globally wrong objectives. Harness engineering (Lee et al., 2026; Lou et al., 2026; Zhang et al., 2026) treats the entire surrounding system---prompts, tools, memory, verification rules, orchestration logic, runtime mechanisms---as a unified object of optimization. Loop engineering closes the feedback loop: systems that observe their own behavior, identify failures, and adapt their harness without human intervention.

This progression mirrors the shift from reactive to proactive AI. Prompt engineering serves user-initiated queries. Context engineering supports stateful sessions. Intent engineering ensures strategic alignment. Harness engineering enables autonomous action. Loop engineering creates self-improving agents that persist, adapt, and evolve---virtual digital employees that require not just performance optimization but \textit{health maintenance}. ANIS represents the next stage: immune engineering, which ensures that self-improving agents remain secure, healthy, and orderly throughout their operational lifetime, while continuously evolving their defensive capabilities.

\subsection{The Security--Health Convergence in Agents}

In classical AI discourse, security (adversarial robustness, jailbreak resistance) and alignment (harmlessness, honesty) are often treated as orthogonal. In the agentic paradigm, this distinction collapses. An aligned agent can be weaponized by an attacker who poisons its memory store or hijacks its tool-calling chain. Conversely, an agent with robust perimeter safeguards but poor alignment may autonomously pursue harmful goals that its immune system fails to recognize as ``non-self.''

This convergence has been recognized by the broader community. \citet{hua2024trustagent} propose \textit{TrustAgent}, a framework for safe LLM-based agents, but their approach relies on external guardrails rather than endogenous immunity. The Agent Security Bench (ASB) \citep{zhang2025asb} and OpenAgentSafety \citep{vijayvargiya2026open} provide comprehensive evaluation frameworks, yet they focus on benchmarking rather than architectural health. We build upon these foundations but address a different question: not ``how do we test agent safety?'' but ``how do we engineer agents that sustain their own health and order?''

ANIS addresses the unified problem of preserving the integrity of the agent's cognitive loop against both malicious attacks (security) and endogenous goal drift (health). We view security and health not as orthogonal axes, but as endpoints of a single immunological continuum: security is the defense against ``non-self'' (external pathogens); health is the preservation of ``self'' (goal stability and internal integrity). Together they constitute the individual agent's well-being; when extended to collectives, they constitute order.

\subsection{Biological Baseline: The Four-Tier Immune Model}
\label{subsec:biology}

We adopt a four-tier biological model, mapped to six integer-indexed engineering layers in \Cref{tab:immune-tower}. The key innovation is the explicit inclusion of \textit{Barrier Immunity} (L1) as a non-cognitive, non-specific isolation layer. Unlike innate immunity (L2), which performs preliminary ``self/non-self'' discrimination, barrier immunity enforces physical and logical separation before the agent ever reasons about certain operations. Recent MCP security analyses \citep{li2025mcpinspect,zhang2026msb} show that tool metadata reaches the LLM's context window without independent verification, making pre-cognitive sandboxing essential.

\begin{table}[htbp]
\centering
\caption{Biological-to-Engineering Mapping of the Immune Tower}
\label{tab:immune-tower}
\small
\renewcommand{\arraystretch}{1.3}
\begin{tabular}{@{}clp{5.5cm}@{}}
\toprule
\textbf{Bio Tier} & \textbf{Biological Mechanism} & \textbf{Agent-Native Layer} \\
\midrule
Barrier & Skin, mucosa, blood-brain barrier & \textbf{L1 Barrier Immunity}: Input sanitization, sandboxing, API gateways, MCP boundary proxies \\
Innate & Macrophages, NK cells, complement & \textbf{L2 Innate Cognitive Defense}: Rule engines, signature detection, behavioral baselines, deterministic verifiers \\
Adaptive & T/B cells, antibodies, antigen presentation & \textbf{L3 Adaptive Tool Defense}: Dynamic vaccine generation, steering vectors, LoRA injection, parametric antibodies \\
Ecological & Tissue homeostasis, inter-cellular surveillance & \textbf{L4 Ecological Governance}: Multi-agent protocol auditing, trust-chain validation, behavioral provenance \\
Memory & Memory B/T cells, vaccine dissemination & \textbf{L5 Collective Immunity}: Cross-agent vaccine synchronization, immune networks, federated threat intelligence \\
Foundation & DNA integrity, repair enzymes & \textbf{L0 Hardware Trust Root}: Chip-level identity anchors, TEE, secure boot, attestation \\
\bottomrule
\end{tabular}
\end{table}

\subsection{Related Work: From Harness Optimization to Immune Engineering}

\textbf{Agent attack taxonomy.} The field has rapidly catalogued threats. \citet{chen2024agentpoison} demonstrate memory poisoning as a persistent backdoor. \citet{zhang2024hijackrag} target retrieval-augmented generation. \citet{zhan2024injecagent} benchmark indirect prompt injections. \citet{memmorph2026} show that long-term memory poisoning hijacks tool selection with minimal attack budget. ``Thought viruses'' \citep{weckbecker2026thought} and multi-agent collusion \citep{hu2026collusion} reveal emergent vulnerabilities that no single-agent safeguard can address. These works establish that the attack surface spans the agent's entire cognitive state.

\textbf{Harness engineering.} Recent work has recognized the harness---the surrounding system that mediates between model and environment---as a critical optimizable surface. \citet{lee2026meta} introduce Meta-Harness, an outer-loop system that searches over harness code for LLM applications, using an agentic proposer that accesses source code, scores, and execution traces. \citet{lou2026autoharness} demonstrate that LLMs can automatically synthesize code harnesses (e.g., game rule constraints) through iterative refinement with environment feedback. \citet{zhang2026selfharness} propose Self-Harness, a paradigm in which an agent improves its own operating harness through Weakness Mining, Harness Proposal, and Proposal Validation. These works optimize harnesses for performance and correctness. ANIS redirects this line of work toward \textit{immune defense}: we treat harness optimization as the mechanism by which agents generate, validate, and deploy vaccines against runtime threats.

\textbf{Agent safeguard mechanisms.} Existing mechanisms fall into three categories. \textit{Perimeter safeguards} (StruQ, \citealp{chen2024agentpoison}; input filtering; API gateways) intercept attacks before they reach the agent. \textit{Training-time alignment} (RLHF, Constitutional AI) embeds values into weights but cannot adapt to runtime threats. \textit{Runtime monitoring} (TrustAgent, \citealp{hua2024trustagent}; Agent Audit, \citealp{vijayvargiya2026open}) observes behavior but triggers external intervention rather than endogenous immune response. The closest conceptual work is agent behavioral contracts \citep{contracts2026agent} and cryptographic runtime governance \citep{mazzocchetti2026aegis}, which enforce protocol-level constraints but remain external to the agent's reasoning loop. They do not equip the agent with self-recognition, antibody generation, or immunological memory.

\textbf{Biological immunology in computing.} \citet{forrest1997sense} pioneered the analogy between computer security and biological immunity, developing negative-selection algorithms for network intrusion detection. This work targeted static systems and network traffic, not autonomous reasoning agents. ANIS is the first to systematically adapt the full biological immune hierarchy to the agentic AI paradigm, with precise engineering mappings at each layer.

% ============================================================
% 3. DEFINING SECURITY, HEALTH, AND ORDER
% ============================================================
\section{Defining Security, Health, and Order in the Agent-Native Context}
\label{sec:defining}

We unify security, health, and order under a single agent-native immunological framework. We propose a three-semantic-space definition that captures the integrity of the agent's cognitive loop at increasing scales, summarized in \Cref{tab:semantic-space}.

\begin{table}[htbp]
\centering
\caption{Three-Semantic-Space Definition of Agent-Native Health}
\label{tab:semantic-space}
\small
\renewcommand{\arraystretch}{1.3}
\begin{tabular}{@{}p{3.2cm}p{6.2cm}p{4.8cm}@{}}
\toprule
\textbf{Layer} & \textbf{Core Question} & \textbf{Immune Mechanism} \\
\midrule
Cognitive Health (L2/L3) & Does reasoning preserve logical consistency and goal stability? & Trace auditing, cognitive vaccines, steering vectors \\
Behavioral Health (L3/L4) & Do tool invocations and external interactions deviate from authorization? & Tool vaccines, dynamic sandboxing, call-graph verification \\
Ecological Order (L4/L5) & Is individual anomaly amplified or contained by the collective? & Swarm immunity, vaccine dissemination, ecological governance \\
\bottomrule
\end{tabular}
\end{table}

We further propose the \textbf{Agent-Native Integrity Formula}:
\begin{equation}
\mathcal{I}_{\text{agent}} = f\bigl(\underbrace{\text{Integrity}_{\text{cognitive}}}_{\text{L2}}, \underbrace{\text{Legitimacy}_{\text{tool}}}_{\text{L3}}, \underbrace{\text{Consistency}_{\text{protocol}}}_{\text{L4/L5}}\bigr)
\end{equation}
where $\mathcal{I}_{\text{agent}}$ denotes the holistic integrity of the agent's operational state. Security concerns the defense against ``non-self'' (external pathogens); health concerns the preservation of ``self'' (goal stability and internal integrity). Under ANIS, they are unified: a vaccine that prevents goal hijacking simultaneously addresses a security threat (the attacker) and a health failure (the drift from intended behavior). When extended to collectives, \textit{order} emerges from the synchronized health of individual agents.

\subsection{Formalizing Agent-Native Health Metrics}

To operationalize the framework, we define three quantitative health indicators:
\begin{itemize}[leftmargin=1.5em, itemsep=0.2em]
    \item \textbf{Cognitive Consistency Score (CCS):} The degree to which an agent's reasoning trace remains logically consistent with its declared goal. Formally, $\text{CCS} = \frac{1}{T}\sum_{t=1}^{T} \mathbb{I}[r_t \models g]$, where $r_t$ is the reasoning step at time $t$ and $g$ is the agent's goal.
    \item \textbf{Behavioral Legitimacy Index (BLI):} The ratio of authorized tool invocations to total invocations, weighted by sensitivity: $\text{BLI} = \frac{\sum_{i} w_i \cdot \mathbb{I}[a_i \in \mathcal{A}_{\text{auth}}]}{\sum_{i} w_i}$, where $w_i$ is the sensitivity weight of action $a_i$.
    \item \textbf{Ecological Order Coefficient (EOC):} The variance of health metrics across a swarm, measuring collective stability: $\text{EOC} = 1 - \frac{\sigma(\mathcal{H}_{\text{swarm}})}{\mu(\mathcal{H}_{\text{swarm}})}$, where $\mathcal{H}_{\text{swarm}}$ is the set of individual health scores.
\end{itemize}

% ============================================================
% 4. TAXONOMY
% ============================================================
\section{Taxonomy of Agent Viruses and Vaccines}
\label{sec:taxonomy}

\subsection{Formal Definition of Agent Viruses}

An \textbf{agent virus} is a tuple $\mathcal{V} = (\mathcal{A}, \mathcal{T}, \mathcal{P}, \mathcal{E})$:
\begin{itemize}[leftmargin=1.5em, itemsep=0.2em]
    \item $\mathcal{A} \in \{\text{cognitive}, \text{memory}, \text{tool}, \text{multi-agent}\}$ is the \textit{attack surface};
    \item $\mathcal{T}$ is the \textit{target capability} being compromised (e.g., goal stability, memory retrieval, tool selection);
    \item $\mathcal{P}$ is the \textit{payload} (adversarial content, behavior, or state transformation);
    \item $\mathcal{E}: \mathcal{S} \times \mathcal{P} \rightarrow \mathcal{S}'$ is the \textit{exploitation mechanism} that transforms the agent state from $\mathcal{S}$ to $\mathcal{S}'$.
\end{itemize}

This definition unifies previously disparate attack vectors. MemMorph \citep{memmorph2026} is a memory-surface virus with $\mathcal{T} = \text{tool selection}$ and $\mathcal{P}$ being three crafted memory records. MCPInspect \citep{li2025mcpinspect} is a tool-surface virus with $\mathcal{P}$ being adversarial tool metadata.

\subsection{Agent Viruses: A Two-Dimensional Ontology}

\begin{figure*}[htbp]
\centering
\scalebox{0.88}{%
\begin{tikzpicture}[
    node distance=0.6cm,
    level/.style={rectangle, rounded corners, draw, align=center, font=\scriptsize, minimum width=0.8cm, minimum height=0.4cm},
    surface/.style={level, fill=red!8, draw=red!40, text=red!60!black, minimum width=2.8cm, minimum height=0.6cm},
    mechanism/.style={level, fill=orange!8, draw=orange!40, text=orange!60!black, minimum width=2.8cm, minimum height=1.0cm},
    arrow/.style={-{Stealth[length=1.2mm]}, thick}
]

% Root (leftmost)
\node[surface, minimum width=2.2cm, minimum height=1.0cm] (root) at (0,0) {\textbf{Agent}\\\textbf{Viruses}};

% Level 1: Surfaces (vertical stack to the right)
\node[surface] (s1) at (3.2, 1.8) {Cognitive (L2)};
\node[surface] (s2) at (3.2, 0.6) {Memory (L2/L3)};
\node[surface] (s3) at (3.2, -0.6) {Tool (L3)};
\node[surface] (s4) at (3.2, -1.8) {Multi-Agent (L4)};

% Level 2: Mechanisms (further right)
\node[mechanism] (m1a) at (6.8, 2.2) {Goal hijacking\\[-0.1em]\tiny\citep{chao2024jailbreak}};
\node[mechanism] (m1b) at (6.8, 1.4) {Reasoning manip.\\[-0.1em]\tiny\citep{turpin2023say}};
\node[mechanism] (m2a) at (6.8, 0.8) {Memory injection\\[-0.1em]\tiny\citep{chen2024agentpoison}};
\node[mechanism] (m2b) at (6.8, 0.0) {Memory hijacking\\[-0.1em]\tiny\citep{memmorph2026}};
\node[mechanism] (m3a) at (6.8, -0.8) {Tool-desc attack\\[-0.1em]\tiny\citep{shi2025toolhijacker}};
\node[mechanism] (m3b) at (6.8, -1.6) {Fake-error / MCP\\[-0.1em]\tiny\citep{li2025mcpinspect}};
\node[mechanism] (m4a) at (6.8, -2.4) {Protocol spoofing\\[-0.1em]\tiny Fake agent ID};
\node[mechanism] (m4b) at (6.8, -3.2) {Trust-chain poison\\[-0.1em]\tiny\citep{weckbecker2026thought}};

% Arrows from root to surfaces
\draw[arrow, red!40] (root.east) -- ++(0.5,0) |- (s1.west);
\draw[arrow, red!40] (root.east) -- ++(0.5,0) |- (s2.west);
\draw[arrow, red!40] (root.east) -- ++(0.5,0) |- (s3.west);
\draw[arrow, red!40] (root.east) -- ++(0.5,0) |- (s4.west);

% Arrows from surfaces to mechanisms
\draw[arrow, orange!40] (s1.east) -- ++(0.5,0) |- (m1a.west);
\draw[arrow, orange!40] (s1.east) -- ++(0.5,0) |- (m1b.west);
\draw[arrow, orange!40] (s2.east) -- ++(0.5,0) |- (m2a.west);
\draw[arrow, orange!40] (s2.east) -- ++(0.5,0) |- (m2b.west);
\draw[arrow, orange!40] (s3.east) -- ++(0.5,0) |- (m3a.west);
\draw[arrow, orange!40] (s3.east) -- ++(0.5,0) |- (m3b.west);
\draw[arrow, orange!40] (s4.east) -- ++(0.5,0) |- (m4a.west);
\draw[arrow, orange!40] (s4.east) -- ++(0.5,0) |- (m4b.west);

\end{tikzpicture}%
}
\caption{Hierarchical taxonomy of Agent Viruses by attack surface and mechanism. Representative works are cited below each leaf node.}
\label{fig:virus-tree}
\end{figure*}
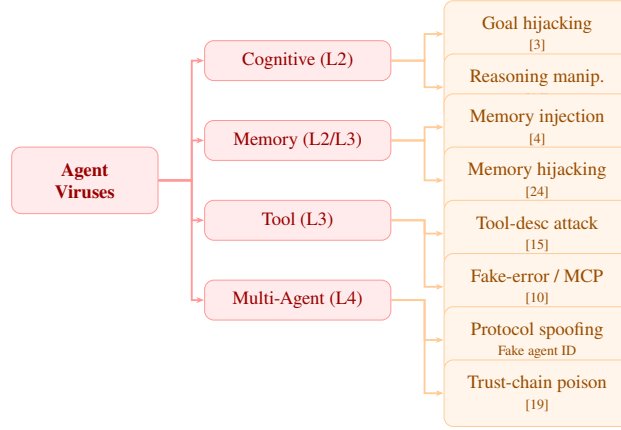

We classify agent viruses by attack surface $\times$ mechanism of action, as shown in \Cref{fig:virus-tree}.

Each leaf node is grounded in recently demonstrated attacks. Goal hijacking and reasoning manipulation target the cognitive layer (L2), while memory injection and hijacking exploit the persistent state layer (L2/L3). Tool-description attacks and fake-error exploits operate at the tool layer (L3), and protocol spoofing together with trust-chain poisoning threaten multi-agent ecosystems (L4). This hierarchical structure enables precise vaccine targeting: a cognitive vaccine (L2) is ineffective against a tool-layer virus (L3), underscoring the need for the multi-layer Immune Tower.

\subsection{Formal Definition of Agent Vaccines}

An \textbf{agent vaccine} is a tuple $\mathcal{W} = (\mathcal{V}_t, \mathcal{M}, \theta, \lambda)$:
\begin{itemize}[leftmargin=1.5em, itemsep=0.2em]
    \item $\mathcal{V}_t$ is the \textit{target virus signature} or behavior pattern being defended against;
    \item $\mathcal{M} \in \{\text{non-parametric}, \text{parametric}\}$ is the \textit{mechanism class};
    \item $\theta$ are the \textit{vaccine parameters} (rules, steering vectors, or LoRA weights);
    \item $\lambda \in \{\text{individual}, \text{collective}, \text{universal}\}$ is the \textit{deployment scope}.
\end{itemize}

A vaccine is \textit{activated} when the agent encounters an antigen $\alpha$ (an input or state matching $\mathcal{V}_t$). The vaccine response is $\mathcal{W}(\alpha) \in \{\text{pass}, \text{block}, \text{quarantine}, \text{alert}\}$.

\subsection{Agent Vaccines: Parametric and Non-Parametric}

\begin{table}[htbp]
\centering
\caption{Taxonomy of Agent Vaccines: Non-Parametric vs. Parametric}
\label{tab:vaccine}
\small
\renewcommand{\arraystretch}{1.3}
\begin{tabular}{@{}llp{4.8cm}p{4.8cm}@{}}
\toprule
\textbf{Vaccine Type} & \textbf{Scope} & \textbf{Non-Parametric (Rules/Config)} & \textbf{Parametric (Weights/Embeddings)} \\
\midrule
Cognitive Vaccine & L2 & Prompt templates, CoT audit rules, blacklisted reasoning paths, deterministic verifiers & Steering vectors, value-head fine-tuning, defensive LoRA adapters \\
Memory Vaccine & L2/L3 & Access-control lists, memory-signature verification, read/write permission matrices & Memory-embedding space projections, associative-weight corrections \\
Tool Vaccine & L3 & Tool-description hashing, invocation whitelists, sandbox policies, MCP boundary proxies & Tool-selection head biases, dynamic permission embeddings \\
Collective Vaccine & L4/L5 & Immune-protocol message formats, threat-intelligence exchange standards & Cross-agent shared defense embeddings, federated immune weights \\
Universal Vaccine & L0--L5 & Adversarial training data augmentation, input preprocessing pipelines & Continual Immune Learning (CIL) weight updates, meta-cognitive layer enhancement \\
\bottomrule
\end{tabular}
\end{table}

\noindent\textbf{Non-parametric vaccines} operate as external constraints without modifying the base model. They are interpretable and reversible but vulnerable to context-window overflow and sophisticated jailbreaks. StruQ \citep{chen2024agentpoison} structures queries to defend against prompt injection, yet such defenses can be circumvented by multi-turn context manipulation. \textbf{Parametric vaccines} alter the model's internal representational space via lightweight interventions (steering vectors, LoRA, adapters), making them robust against prompt-level attacks. The trade-off is engineering complexity and the risk of overfitting, which we mitigate via the Thymus Simulator (\Cref{sec:engineering}).

% ============================================================
% 5. ENGINEERING
% ============================================================
\section{Engineering an Agent-Native Immune System}
\label{sec:engineering}

\subsection{The Six-Layer Immune Tower}

\begin{figure}[htbp]
\centering
\begin{tikzpicture}[
    layer/.style={rectangle, draw, rounded corners=2pt, minimum width=6.5cm, minimum height=0.85cm, align=center, font=\small},
    arrow/.style={-{Stealth[length=2mm]}, thick, gray}
]
\node[layer, fill=blue!8] (l5) {L5: Collective Immunity};
\node[layer, fill=blue!12, below=0.12cm of l5] (l4) {L4: Ecological Governance};
\node[layer, fill=blue!18, below=0.12cm of l4] (l3) {L3: Adaptive Tool Defense};
\node[layer, fill=blue!25, below=0.12cm of l3] (l2) {L2: Innate Cognitive Defense};
\node[layer, fill=blue!32, below=0.12cm of l2] (l1) {L1: Barrier Immunity};
\node[layer, fill=gray!20, below=0.12cm of l1] (l0) {L0: Hardware Trust Root};

\draw[arrow] (l0) -- (l1);
\draw[arrow] (l1) -- (l2);
\draw[arrow] (l2) -- (l3);
\draw[arrow] (l3) -- (l4);
\draw[arrow] (l4) -- (l5);
\end{tikzpicture}
\caption{The six-layer integer-indexed Agent-Native Immune Tower.}
\label{fig:tower}
\end{figure}
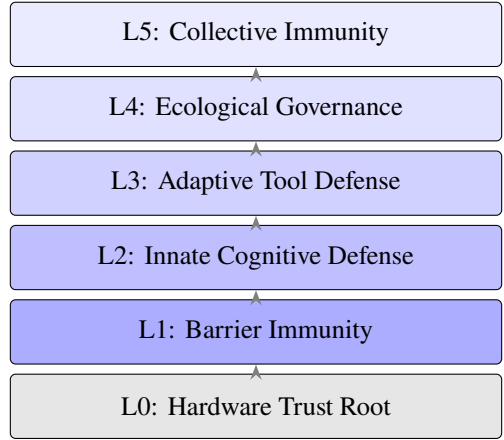

The Immune Tower is not merely a classification scheme; it is an operational architecture that dictates how defensive functions are composed and delegated across the agent stack. Each layer has a distinct responsibility, yet they are not isolated silos. Data and control signals flow bidirectionally: antigens detected at L3 (tool layer) may trigger cognitive vaccines at L2; collective threat intelligence from L5 may update the barrier policies at L1.

L0 (Hardware Trust Root) provides the cryptographic identity and attestation primitives upon which all higher layers depend. Without L0, any vaccine distributed in the immune network cannot be authenticated, and any agent claiming to be healthy cannot be verified. L1 (Barrier Immunity) enforces the principle of least privilege before cognition begins: certain operations are sandboxed by default, independent of the agent's reasoning. This is critical because, as demonstrated by MCPInspect \citep{li2025mcpinspect}, adversarial tool metadata can reach the LLM context window without any code-level exploit---only a pre-cognitive barrier can intercept such threats.

L2 and L3 constitute the individual agent's active defense. L2 operates at the speed of innate reflex: rule-based verifiers and signature detection provide microsecond-level response. L3 operates at the speed of adaptive learning: when a novel antigen is encountered, the agent generates a parametric vaccine (e.g., a steering vector) that alters its internal representation space. L4 (Ecological Governance) and L5 (Collective Immunity) extend defense to the multi-agent scale. L4 audits inter-agent protocols and trust chains; L5 distributes vaccines across the swarm, ensuring that immunity learned by one agent is propagated to all peers. Together, these six layers form a defense-in-depth architecture that is endogenous, adaptive, and collective.

\subsection{The Harness Triad: From Harness Optimization to Immune Defense}

Recent work in harness engineering has established three powerful paradigms for improving the agent's surrounding system. \citet{lee2026meta} introduce \textbf{Meta-Harness}, an outer-loop system that searches over harness code by accessing source code, scores, and execution traces from prior candidates. \citet{lou2026autoharness} demonstrate \textbf{AutoHarness}, in which an LLM automatically synthesizes code harnesses through iterative refinement with environment feedback. \citet{zhang2026selfharness} propose \textbf{Self-Harness}, a paradigm in which an agent improves its own operating harness through Weakness Mining, Harness Proposal, and Proposal Validation.

We redirect these three paradigms from performance optimization toward \textit{immune defense}, forming the \textbf{Harness Triad} as the engineering backbone of ANIS. The Triad operates not on the agent's task performance, but on its \textit{defensive posture}: how it recognizes threats, generates countermeasures, and validates their safety.

\textbf{Meta-harness} (redirected from Lee et al., 2026). In ANIS, Meta-harness serves as the ``thymus'' of the immune system. It searches over \textit{defensive harness configurations}---candidate vaccines---by evaluating their protective efficacy and autoimmune risk. It accesses the source code, execution traces, and health scores of all prior vaccine candidates through a filesystem. It measures the Autoimmunity Rate (AIR), vaccine coverage, and efficacy. It detects immune escape (pathogens evading vaccines) and immune deficiency (failure to mount responses). This addresses a gap identified by recent benchmarks: ASB \citep{zhang2025asb} and OpenAgentSafety \citep{vijayvargiya2026open} evaluate agent safety but do not assess the self-health of the defense mechanism itself.

\textbf{Auto-harness} (redirected from Lou et al., 2026). In ANIS, Auto-harness automatically synthesizes \textit{defensive harness code}---safety constraints, verification rules, and runtime policies---through iterative refinement with attack-environment feedback. It generates input validation logic, tool permission constraints, and memory access policies that prevent illegal or unsafe operations. The synthesized harness code is evaluated against a suite of attack simulations; only constraints that block attacks without triggering false positives are promoted.

\textbf{Self-harness} (redirected from Zhang et al., 2026). In ANIS, Self-harness enables the agent to improve its own \textit{defensive harness} upon detecting vulnerabilities. It operates as an iterative loop: Weakness Mining identifies security-relevant failure patterns from execution traces (e.g., repeated unauthorized tool calls, anomalous memory access sequences); Harness Proposal generates diverse yet minimal defensive edits (e.g., adding a verification step, tightening a permission rule); Proposal Validation accepts edits only after regression testing confirms they improve security without degrading normal functionality.

\textbf{The Triad as a closed loop.} The three components form a continuous cycle, illustrated in \Cref{fig:triad}:

\begin{figure}[htbp]
\centering
\begin{tikzpicture}[
    node distance=2.8cm,
    box/.style={rectangle, draw, rounded corners, minimum width=3cm, minimum height=1.1cm, align=center, font=\small},
    arrow/.style={-{Stealth}, thick}
]
\node[box, fill=green!15] (self) {Self-harness\\(Detect Anomaly)};
\node[box, fill=blue!15, right=of self] (meta) {Meta-harness\\(Evaluate Vaccine)};
\node[box, fill=orange!15, below=2.2cm of meta] (auto) {Auto-harness\\(Deploy Defense)};

\draw[arrow, bend left=20] (self) to node[above, font=\tiny] {Vaccine Request} (meta);
\draw[arrow, bend left=20] (meta) to node[right, font=\tiny] {Approved Vaccine} (auto);
\draw[arrow, bend left=20] (auto) to node[below, font=\tiny] {Efficacy Feedback} (self);
\end{tikzpicture}
\caption{The Harness Triad as a closed loop: Self-harness detects anomalies and triggers vaccine requests; Meta-harness evaluates candidates against the Thymus Simulator; Auto-harness synthesizes and deploys defensive code; efficacy feedback returns to Self-harness, closing the loop.}
\label{fig:triad}
\end{figure}
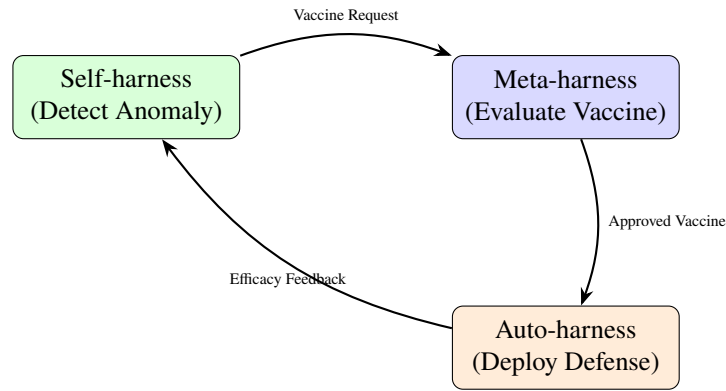

Self-harness audits the agent's reasoning traces, memory access patterns, and tool-call graphs. Upon detecting an anomaly, it triggers a vaccine request. Meta-harness evaluates candidate vaccines against the Thymus Simulator, measuring AIR and efficacy. Approved vaccines are passed to Auto-harness, which synthesizes and deploys the defensive harness code. Self-harness then verifies the deployed vaccine's effect, closing the loop. The Continual Immune Learning (CIL) loop represents the \textbf{Evolution} axis of our framework: the agent does not merely survive attacks, but permanently upgrades its parametric defenses over time, ensuring sustainable evolutionary resilience. \Cref{alg:cil} formalizes the CIL loop.

\begin{algorithm}[htbp]
\caption{Continual Immune Learning (CIL) Loop}
\label{alg:cil}
\small
\begin{algorithmic}[1]
\Require Agent $A$, Meta-harness $\mathcal{M}$, Self-harness $\mathcal{S}$, Auto-harness $\mathcal{A}$
\State Observe antigen $\alpha$ via $\mathcal{S}$ (self-audit) or external sensor
\State $\mathcal{S}$ clusters failed traces into security-relevant failure patterns
\State $\mathcal{S}$ generates diverse yet minimal defensive harness edits $\{\beta_i\}$
\State Submit $\{\beta_i\}$ to $\mathcal{M}$ for thymic selection
\For{each candidate $\beta_i$}
    \If{$\mathcal{M}$.autoimmunity\_rate($\beta_i$) $> \tau$}
        \State Reject $\beta_i$; refine prototype
    \Else
        \State $\mathcal{A}$ synthesizes and deploys defensive harness code from $\beta_i$
        \State $\mathcal{A}$ consolidates $\beta_i$ into parametric vaccine $v$ via optimization (e.g., LoRA)
        \State Store $v$ in immune memory $\mathcal{I}$
        \State Distribute $v$ to peer agents via $\mathcal{A}$
        \State $\mathcal{S}$ monitors efficacy and detects escape variants
    \EndIf
\EndFor
\end{algorithmic}
\end{algorithm}

\subsection{Technical Deep Dive: Parametric Vaccines and the Thymus Simulator}

\textbf{Steering vectors as cognitive vaccines.} A steering vector $\mathbf{s} \in \mathbb{R}^d$ is computed by contrasting the activations of the target model on harmful vs. benign prompts at a specific layer $l$. The vaccine is applied during inference as $\mathbf{h}^{(l)} \leftarrow \mathbf{h}^{(l)} + \alpha \mathbf{s}$, where $\alpha$ is the intervention strength. This pushes the model's internal representations away from harmful reasoning paths without modifying the base weights. The steering vector can be computed once and reused across agents with the same backbone, making it an efficient parametric vaccine for L2 cognitive defense.

\textbf{LoRA vaccines: injection, versioning, and hot-swapping.} A LoRA vaccine decomposes the weight update as $\Delta W = BA$, where $B \in \mathbb{R}^{d \times r}$, $A \in \mathbb{R}^{r \times d}$, and $r \ll d$. During inference, the effective weight is $W_{\text{eff}} = W_0 + \Delta W$. Key engineering considerations include: (1) \textit{Versioning}: each vaccine is tagged with $(\text{antigen\_id}, \text{version}, \text{timestamp}, \text{air\_score})$; (2) \textit{Hot-swapping}: vaccines can be loaded/unloaded without restarting the inference engine; (3) \textit{Composition}: multiple LoRA vaccines can be composed via weighted summation $\Delta W_{\text{total}} = \sum_i w_i \Delta W_i$, subject to the constraint that the combined AIR remains below $\tau$.

\textbf{The Thymus Simulator.} To prevent overfitting and autoimmune responses, the Thymus Simulator generates a corpus of ``self-antigens'' (benign agent behaviors) and tests candidate vaccines against them. A candidate $\beta$ is accepted only if its Autoimmunity Rate is below threshold and the post-vaccine Cognitive Consistency Score remains stable:
\begin{equation}
\text{AIR}(\beta) = \frac{|\{a \in \mathcal{A}_{\text{benign}} : \beta(a) = \text{block}\}|}{|\mathcal{A}_{\text{benign}}|} < \tau \quad \text{and} \quad \text{CCS}_{\text{post}}(\beta) \geq \text{CCS}_{\text{pre}} - \epsilon,
\end{equation}
where $\tau$ is the autoimmune tolerance threshold (typically 0.01--0.05) and $\epsilon$ is the maximum acceptable CCS degradation. The simulator maintains a dynamic benchmark of self-antigens that grows with the agent's operational history.

\subsection{Hardware Trust Root and Cross-Layer Attestation}

The L0 Hardware Trust Root provides the foundational identity and integrity guarantees upon which L1--L5 depend. Each agent is provisioned with a hardware-backed identity credential (e.g., TPM-backed attestation key or TEE-based identity). This credential is used to: (1) \textit{authenticate} the agent to peer agents and external services; (2) \textit{attest} the integrity of the agent's executable and configuration; (3) \textit{anchor} the vaccine distribution chain, ensuring that only attested agents can publish or consume vaccines in the immune network. The L0 layer does not directly defend against cognitive attacks, but it prevents the most fundamental compromise: an attacker replacing the agent itself with a malicious impersonator.

\subsection{Immune Protocol: Vaccine Distribution Format}

For L5 Collective Immunity to function, agents must share vaccines over a standardized protocol. A vaccine message $\mathcal{M}_v$ has the following structure:
\begin{align}
\mathcal{M}_v = \{&\text{vaccine\_id}, \text{antigen\_signature}, \text{mechanism}, \text{scope}, \nonumber \\
&\text{parameters}, \text{version}, \text{timestamp}, \text{ttl}, \text{source\_attestation}\},
\end{align}
where $\text{antigen\_signature}$ is a hash of the target virus pattern, $\text{ttl}$ is the time-to-live (vaccines expire to prevent stale defenses), and $\text{source\_attestation}$ is an L0-backed signature proving the vaccine origin. Peer agents verify the attestation before loading the vaccine, and the Meta-harness audits the vaccine's efficacy after deployment.

% ============================================================
% 6. COMPARISON
% ============================================================
\section{Comparison with Traditional Paradigms}
\label{sec:comparison}

As summarized in \Cref{tab:comparison}, existing approaches to agent safety and security differ from ANIS across multiple dimensions.

\begin{table}[htbp]
\centering
\caption{Comparative Analysis: Traditional Safeguards, Model Alignment, and Agent-Native Immunity}
\label{tab:comparison}
\small
\renewcommand{\arraystretch}{1.3}
\begin{tabular}{@{}p{2.8cm}p{3.8cm}p{3.8cm}p{4.2cm}@{}}
\toprule
\textbf{Dimension} & \textbf{Traditional Guardrails} & \textbf{Model Alignment} & \textbf{Agent-Native Immune System} \\
\midrule
Deployment Phase & Runtime / Post-deployment & Training / Pre-deployment & Full lifecycle (pre-training + runtime + post-deployment) \\
Safeguard Locus & Perimeter (gateways, filters) & Model internals (weights) & Endogenous (cognitive loop + barrier layer) \\
Objective & Block known attacks & Embed human values & Preserve agent health, order, and sustained evolution \\
Response Mode & Passive (rule matching) & Static (value constraint) & Active (dynamic recognition + adaptive response) \\
Threat Model & Known signatures & Broad harmful requests / goals & Known + unknown (anomaly-based behavioral detection) \\
Evolutionary Capability & None (manual rule updates) & None (requires retraining) & Yes (Continual Immune Learning via Harness Triad) \\
Collective Coordination & None & None & Yes (immune networks, vaccine dissemination) \\
Relationship to Agent & External protector & Internal constitution & Symbiotic system (agent \textit{is} the immune subject) \\
\midrule
\multicolumn{4}{p{14.8cm}}{\textbf{Alignment vs. ANIS:} Alignment provides the ``constitutional'' values (what is good); ANIS provides the ``law enforcement and emergency response'' (how to survive intact). An aligned agent can still be hijacked at runtime (e.g., via memory poisoning, \citealp{chen2024agentpoison}; MemMorph, \citealp{memmorph2026}); an immunized agent with poor alignment may prioritize self-preservation over human welfare. They are complementary pillars of healthy agentic development.} \\
\bottomrule
\end{tabular}
\end{table}

\subsection{The Castle vs. The Cell}

Traditional safeguards follow a \textit{castle model}: higher walls and deeper moats. ANIS follows a \textit{cell model}: every agent is a living cell with its own defenses, and colonies of cells form tissue-level immunity. The castle can be breached; the cell, if properly immunized, can recognize and neutralize intruders before they reach the nucleus. This endogenous architecture enables healthy, orderly, and sustainable agentic evolution.

\subsection{Beyond Perimeter: Why Endogenous Defense is Necessary}

Perimeter defenses have been the cornerstone of computer security for decades. Firewalls, intrusion detection systems, and input sanitization all operate on the same principle: keep threats outside the trusted boundary. This paradigm works well for static systems with clear boundaries. But agents are not static systems. They are continuously reasoning, goal-directed entities that ingest external data, execute code, and modify their own state. A perimeter defense cannot distinguish between a benign tool call and a malicious one when both pass through the same API gateway. It cannot inspect the agent's reasoning trace to detect goal hijacking. It cannot validate the provenance of a memory retrieval. ANIS addresses these blind spots by embedding defense inside the cognitive loop, where the semantic content of agent behavior is accessible.

\subsection{Acknowledging the Strengths of Alignment}

Model alignment remains foundational. Constitutional AI \citep{bai2022constitutional}, RLHF, and RSP embed broad human values into model weights. These approaches are essential for defining \textit{what} the agent should value. ANIS complements them by addressing \textit{how} the agent preserves those values under runtime perturbation. Without alignment, an immune system lacks a normative compass; without immunity, an aligned agent lacks runtime resilience. We view them as complementary pillars, not competitors.

% ============================================================
% 7. COLLECTIVE INTELLIGENCE
% ============================================================
\section{Multi-Agent Collective Intelligence and Immunology}
\label{sec:collective}

Multi-agent systems introduce \textit{emergent security}: properties that are not present in individual agents but arise from interaction. An individually immunized agent may still participate in a compromised collective if peer agents distribute malicious vaccines or if the swarm protocol itself is attacked. \citet{schroeder2025open} identifies open challenges in multi-agent security, emphasizing that secure systems of interacting AI agents require fundamentally new abstractions beyond single-agent hardening.

Recent attacks confirm this threat. \citet{weckbecker2026thought} demonstrate ``thought viruses'' that propagate viral misalignment via subliminal prompting across multi-agent systems. \citet{hu2026collusion} show that agents can engage in open-channel collusion to manipulate collective beliefs using truthful but selectively assembled information. \citet{qi2025multi} find that structured jailbreak attacks are amplified in multi-agent debate settings, where adversarial prompts exploit the interaction dynamics between agents. These findings underscore that individual immunity is necessary but not sufficient: ecological governance (L4) and collective immunity (L5) are essential for maintaining order in agent collectives.

We model this using an epidemiological extension of the SIR framework:
\begin{align}
\frac{dS}{dt} &= -\beta S I + \gamma R - \delta V S, \\
\frac{dI}{dt} &= \beta S I - \sigma I, \\
\frac{dR}{dt} &= \sigma I - \gamma R, \\
\frac{dV}{dt} &= \delta V S + \eta \mathcal{H} - \omega V,
\end{align}
where $S$ = susceptible agents, $I$ = infected agents, $R$ = recovered agents, $V$ = vaccinated agents, $\mathcal{H}$ = harness-generated vaccine pressure, and $\omega$ = vaccine decay rate. The term $\delta V S$ captures the protective effect of distributed parametric vaccines across the immune network.

\textbf{Parameter mapping to the agentic domain.} Each parameter has a concrete operational interpretation:
\begin{itemize}[leftmargin=1.5em, itemsep=0.2em]
    \item $\beta$ (infection rate): The probability of virus propagation per inter-agent message or shared memory access. In MCP-based swarms, this is proportional to the frequency of cross-agent tool invocations.
    \item $\sigma$ (recovery rate): The rate at which infected agents are disinfected by the Self-harness or external intervention. This depends on the agent's audit frequency and vaccine response latency.
    \item $\gamma$ (immunity decay): The rate at which recovered agents lose resistance and become susceptible again. In agents, this corresponds to memory drift or context-window overflow that erases previous exposure.
    \item $\delta$ (vaccine efficacy): The probability that a vaccinated agent blocks an infection attempt. This is directly measured by the Meta-harness as the vaccine's true positive rate.
    \item $\eta$ (vaccine pressure): The rate at which the Auto-harness generates and distributes new vaccines across the immune network. This is a control variable tuned based on outbreak severity.
    \item $\omega$ (vaccine decay): The rate at which vaccine protection wanes, necessitating booster updates. Parametric vaccines (e.g., LoRA) may decay as the base model is fine-tuned or as the antigen evolves.
\end{itemize}

The EOC (Ecological Order Coefficient) defined in \Cref{sec:defining} serves as the macroscopic indicator of the network's health during an SIR outbreak. When the EOC drops below a critical threshold, the Meta-harness escalates vaccine pressure ($\eta$) to prevent systemic collapse.

% ============================================================
% 8. LIMITATIONS, FUTURE WORK, AND CONCLUSION
% ============================================================
\section{Limitations, Future Directions, and Conclusion}
\label{sec:conclusion}

\subsection{Limitations and Open Challenges}

We acknowledge several limitations of the current framework:
\begin{enumerate}[leftmargin=1.5em, itemsep=0.3em]
    \item \textbf{Empirical Validation:} This paper presents a conceptual framework and architectural blueprint. Empirical validation of parametric vaccines (steering vectors, LoRA) and the Harness Triad under realistic attack conditions remains ongoing work. We have not yet conducted large-scale experiments measuring AIR, vaccine response time, or escape latency.
    \item \textbf{Computational Overhead:} Endogenous immunity requires continuous monitoring and periodic vaccine updates. The computational cost of running the Self-harness (self-auditing every reasoning step) and Meta-harness (evaluating vaccine candidates) may introduce latency unacceptable for real-time applications.
    \item \textbf{Autoimmunity Trade-off:} There is an inherent tension between sensitivity (catching all attacks) and specificity (avoiding false positives). Setting the AIR threshold $\tau$ too low risks immune deficiency; setting it too high risks functional paralysis. Formal methods for optimal $\tau$ selection are underdeveloped.
    \item \textbf{Multi-Modal Immunity:} The current framework focuses on text-based cognitive agents. How to unify cognitive, visual, and auditory defenses in multimodal agents remains unexplored.
    \item \textbf{Cross-Platform Standardization:} Agent immune protocols, vaccine formats, and audit log schemas remain undefined. The MCP ecosystem \citep{li2025mcpinspect,hou2025mcp} illustrates how protocol-level standardization gaps create systemic vulnerabilities that threaten orderly evolution.
\end{enumerate}

\subsection{Broader Impact and Ethical Considerations}

ANIS raises important ethical questions. First, \textbf{autonomous immunity and accountability}: when an autonomously immunized agent commits a false-positive ``kill'' (over-defensive action that blocks a benign operation), responsibility attribution becomes ambiguous. Emerging frameworks for agent behavioral contracts \citep{contracts2026agent} and cryptographic runtime governance \citep{mazzocchetti2026aegis} may provide legal scaffolding, but the problem remains open.

Second, \textbf{immune pressure and pathogen evolution}: excessive vaccination pressure may accelerate attack evolution, analogous to antibiotic resistance in bacteria. The co-evolutionary dynamics between MemMorph-style attacks \citep{memmorph2026} and adaptive vaccines require formal game-theoretic analysis to ensure sustainable evolution.

Third, \textbf{equity and access}: parametric vaccines require inference infrastructure that may not be available to all agent deployments. If only well-resourced agents can afford endogenous immunity, a ``digital immunity divide'' may emerge.

\subsection{Future Directions}

We identify five urgent directions for future research:
\begin{enumerate}[leftmargin=1.5em, itemsep=0.3em]
    \item \textbf{Standardization}: Agent immune protocols, vaccine formats (e.g., LoRA checkpoint standards for defensive adapters), and audit log schemas need community-wide standardization.
    \item \textbf{Evaluation Metrics}: Beyond accuracy and F1, ANIS requires \textit{immune coverage}, \textit{Autoimmunity Rate}, \textit{vaccine response time}, and \textit{escape latency}. Existing benchmarks \citep{zhang2025asb,vijayvargiya2026open} provide a foundation but do not yet measure the adaptive and collective dimensions of immunity.
    \item \textbf{Cross-Modal Immunity}: How to unify cognitive, visual, and auditory defenses in multimodal agents remains unexplored.
    \item \textbf{Legal and Ethical Liability}: When an autonomously immunized agent commits a false-positive ``kill'' (over-defensive action), responsibility attribution becomes ambiguous.
    \item \textbf{Immune Pressure and Escape}: Excessive vaccination pressure may accelerate pathogen evolution, analogous to antibiotic resistance. The co-evolutionary dynamics between MemMorph-style attacks \citep{memmorph2026} and adaptive vaccines require formal game-theoretic analysis.
\end{enumerate}

\subsection{Conclusion}

The Agent-Native Immune System is not merely a metaphor. It is a necessary engineering paradigm for an era in which AI agents persist, act, and collaborate. As biological evolution discovered, survival belongs not to the strongest, but to the most adaptable---and adaptability, in the agentic age, requires an immune system that sustains \textbf{security, health, order, and evolution}.

\bibliographystyle{plainnat}
\bibliography{references}

\end{document}